\documentclass[conference]{IEEEtran}
\IEEEoverridecommandlockouts
\usepackage{cite}
\usepackage{amsmath,amssymb,amsfonts}
\usepackage{algorithmic}
\usepackage{graphicx}
\usepackage{textcomp}
\usepackage{xcolor}
\def\BibTeX{{\rm B\kern-.05em{\sc i\kern-.025em b}\kern-.08em
    T\kern-.1667em\lower.7ex\hbox{E}\kern-.125emX}}
\usepackage[table]{xcolor}
\usepackage{soul}

\begin{document}

\title{CRISP: Pre-LLM Yet Text-Driven Visual Token Pruning for Efficient LVLM Inference
% {\footnotesize \textsuperscript{*}Note: Sub-titles are not captured for https://ieeexplore.ieee.org  and
% should not be used}
% \thanks{Identify applicable funding agency here. If none, delete this.}
}

\author{
Xu Li$^{1}$, Yi Zheng$^{1}$, Mengyang Zhao$^{1}$, Yuxuan Liang$^{1}$, 
Zhe Liu$^{1}$, Rui Zhu$^{1}$, Xiaolei Chen$^{1}$, 
\\Wei Zhou$^{2}$, Baoquan Zhao$^{3}$, Juncen Guo$^{4\dagger}$\\
$^{1}$College of Computer Science and Artificial Intelligence, Fudan University\\
$^{2}$School of Computer Science and Informatics, Cardiff University\\
$^{3}$School of Artificial Intelligence, Sun Yat-sen University\\
$^{4}$College of Intelligent Robotics and Advanced Manufacturing, Fudan University\\
{\tt\small \{xu\_li23, 23210240407, myzhao20, yxliang25, 24210240239, rzhu24, chenxl23, guojc23\}@m.fudan.edu.cn}\\
{\tt\small zhouw26@cardiff.ac.uk, zhaobaoquan@mail.sysu.edu.cn}\\
{\small $^{\dagger}$Corresponding author}
}

\maketitle

\begin{abstract}
Large Vision-Language Models (LVLMs) typically require processing hundreds to thousands of visual tokens, leading to substantial inference overhead. Existing visual token pruning methods either operate before the LLM using text-agnostic heuristics or prune inside the LLM at the cost of efficiency and noisy cross-modal attention. To address these limitations, we propose CRISP, a pre-LLM yet text-driven visual token pruning framework that preserves both instruction-relevant evidence and essential scene context. CRISP works in a two-stage pipeline: stage-1 first identifies text-aligned visual tokens, and then stage-2 enhances contextual completeness through semantic diversity. Extensive experiments on LLaVA-1.5 and LLaVA-NeXT demonstrate that CRISP achieves superior performance retention under aggressive pruning ratios, maintaining up to 99.5\% accuracy while reducing inference cost and latency by more than 2$\times$. CRISP serves as a practical solution for efficient LVLM inference, especially in resource-constrained scenarios.
\end{abstract}


\section{Introduction}
% As an emerging family of generative models, Large Vision-Language Models (LVLMs) combine vision encoders with Large Language Models (LLMs) to achieve general-purpose capabilities across vision-language tasks \cite{lvlm-survey}. However, these models typically encode an image as a long sequence of visual tokens that are fully processed by the LLM, incurring substantial computational overhead. This heavy token load poses a fundamental barrier to efficient inference, limiting the practicality of LVLMs in latency-sensitive applications.

Large Vision-Language Models (LVLMs) combine vision encoders with Large Language Models (LLMs) to support general-purpose vision-language tasks \cite{lvlm-survey}. However, they usually encode images as long visual token sequences that are fully processed by the LLM, causing substantial computational overhead and hindering deployment in latency-sensitive or resource-constrained scenarios.

Recent studies have shown that the visual tokens in LVLMs contain substantial redundancy \cite{fastv, fastervlm, dart}, making training-free visual token pruning an increasingly active research direction. Depending on where the pruning is performed, existing approaches can be broadly categorized into two types: pre-LLM pruning and intra-LLM pruning. Pre-LLM methods discard unnecessary visual tokens before they are passed into the language model, typically by using CLS attention from the vision encoder as importance scores \cite{visionzip}, or by eliminating tokens that exhibit high feature similarity\cite{divprune}. In contrast, intra-LLM methods prune visual tokens after they enter the LLM, often leveraging cross-modal attention (i.e., the attention scores assigned by textual tokens to visual tokens) to remove those that are least relevant to the current textual input \cite{sparsevlm}. 

\begin{figure}[t]
\centering
\includegraphics[width=0.85\columnwidth]{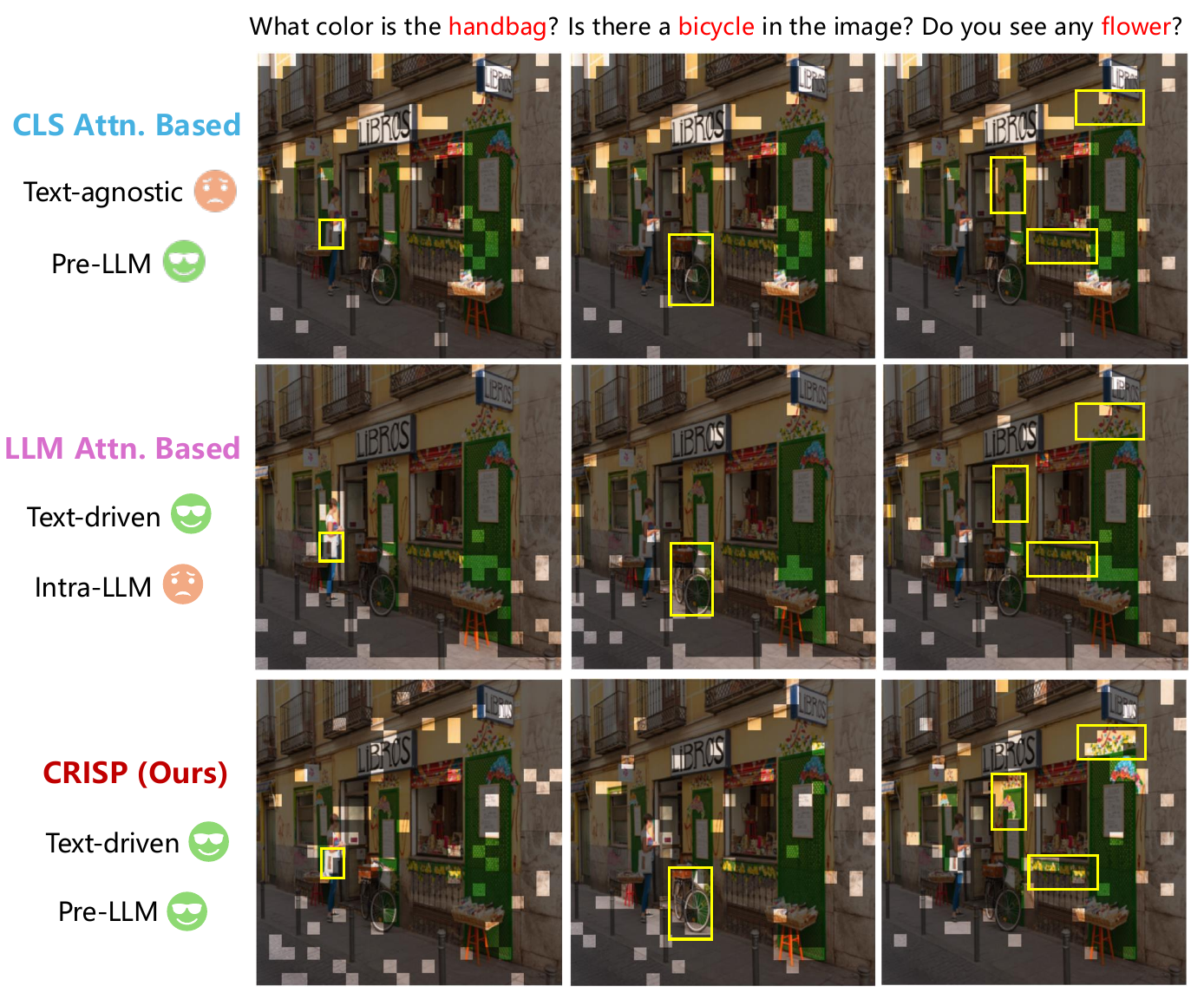}
\caption{Comparison of visual token pruning behaviors under different textual queries on the same image. Masked patches indicate the pruned tokens. Answer-relevant regions are highlighted with yellow bounding boxes.}
\label{fig1}
\end{figure}

\begin{figure}[t]
\centering
\includegraphics[width=0.8\columnwidth]{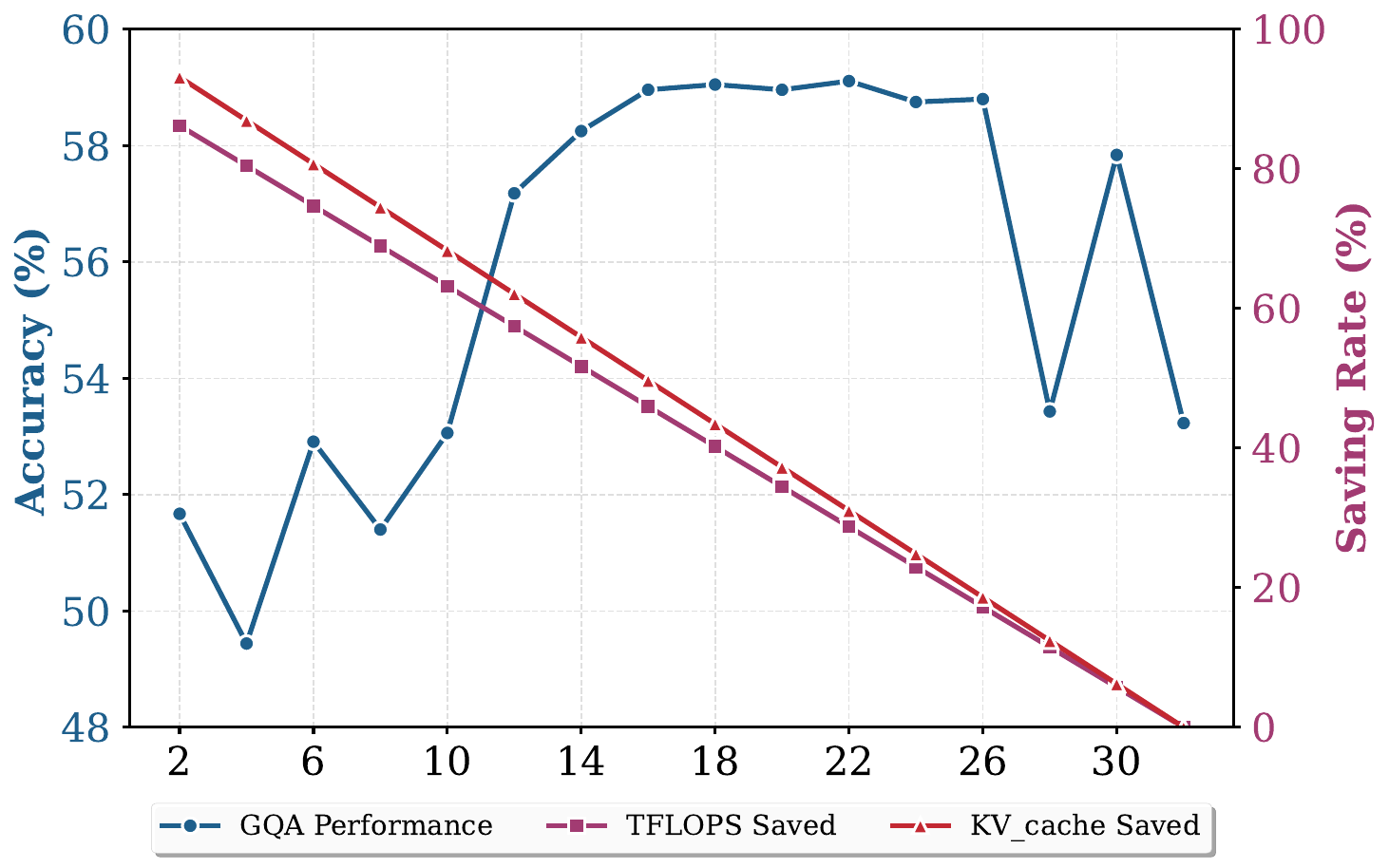}
\caption{Accuracy–efficiency trade-off when pruning at different LLM layers.}
\label{fig2}
\end{figure}

Despite the effectiveness of these methods, they have certain limitations. Pre-LLM pruning is typically text-agnostic: for a given image, its pruning results remain unchanged regardless of the textual query. As shown in Fig.~\ref{fig1}, when querying about different elements of an image, the CLS attention-based method consistently retains the same token set. This behavior resembles a “look-first-then-ask’’ strategy, requiring the model to preserve the important details before knowing the task. Cognitive studies have shown that such a “see-then-read’’ paradigm imposes higher cognitive load and yields worse task performance compared to the more natural “read-then-see’’ process \cite{cognition1,cognition2}. 
In contrast, intra-LLM pruning is text-driven but suffers from a significant performance–efficiency trade-off. As shown in Fig.~\ref{fig2}, pruning at early LLM layers severely degrades accuracy due to the well-known positional bias and attention dispersion present in shallow layers' cross-modal attention \cite{vispruner}. While pruning at deeper layers alleviates these issues, it requires all visual tokens to pass through many LLM blocks, diminishing the computational savings from pruning. Moreover, extracting attention weights from the LLM disrupts FlashAttention\cite{flashattention}, further harming inference efficiency.
These limitations highlight the need for a new approach that can \textbf{(i)} perform visual token pruning before the LLM but \textbf{(ii)} remain text-driven, ensuring that task-relevant tokens can be effectively retained.

% Motivated by these limitations, we propose \textbf{CRISP}, a \textbf{CRI}tical-first and \textbf{S}emantically-covered \textbf{P}re-LLM token pruning framework that achieves text-driven visual token selection entirely before the LLM. First, it identifies a compact set of critical visual tokens by computing cross-modal relevance between the visual tokens and key textual elements in a shared semantic space, ensuring that token selection is strongly aligned with the query intent. Second, it performs semantic coverage via a diversity-driven completion step that selects additional tokens to recover essential scene context beyond the query-aligned evidence. This design enables CRISP to remain both text-aware and computationally efficient without accessing any LLM attention.
% Despite its simplicity and training-free nature, CRISP delivers strong empirical gains. On LLaVA-1.5-7B \cite{llava1.5}, it preserves 99.5\% of the overall performance by retaining only 22.2\% visual tokens. On LLaVA-NeXT-7B \cite{llavanext}, it keeps merely 11.1\% of the tokens while still maintaining 96.9\% of the performance. These results demonstrate that CRISP offers a highly effective and practical solution for pre-LLM visual token pruning in modern LVLMs.

Motivated by these limitations, we propose \textbf{CRISP}, a \textbf{CRI}tical-first and \textbf{S}emantically-covered \textbf{P}re-LLM token pruning framework for efficient LVLM inference. CRISP first identifies critical visual tokens by measuring cross-modal relevance between visual tokens and key textual elements in a shared semantic space, ensuring alignment with the query intent. It then applies a diversity-driven completion step to recover essential scene context beyond the query-related evidence. This design enables CRISP to remain both text-aware and computationally efficient without accessing any LLM attention.
Despite being simple and training-free, CRISP achieves strong empirical results. On LLaVA-1.5-7B, it preserves 99.5\% of overall performance while retaining only 22.2\% of visual tokens. On LLaVA-NeXT-7B, it keeps merely 11.1\% of the tokens while still maintaining 96.9\% of the performance. These results demonstrate that CRISP offers a highly effective and practical solution for pre-LLM visual token pruning in modern LVLMs.

\section{Related Work}
\subsection{Large Vision-Language Models}
Large Vision-Language Models (LVLMs) typically consist of a vision encoder, a projector and an LLM \cite{llava1.5}. The vision encoder first extracts dense visual features from an input image, which are then mapped into the LLM’s input embedding space through the projector. The resulting visual tokens are concatenated with textual tokens along the sequence dimension and jointly processed by the LLM to enable multimodal understanding and text generation. While effective, this design often produces hundreds or even thousands of visual tokens, especially under high resolution inputs \cite{llavanext}, leading to substantial inference overhead. 
Since most visual tokens are only marginally relevant to a given query \cite{fastv, avl}, reducing the number of visual tokens is critical to improving the efficiency and scalability of LVLMs.

\subsection{Visual Token Pruning in LVLMs}
A variety of visual token pruning methods have been proposed for LVLMs. The first class of methods performs pruning before the LLM by relying solely on visual signals. While FasterVLM \cite{fastervlm}, LLaVA-Prumerge \cite{prumerge}, and HiRED \cite{hired} leverage CLS attention from the vision encoder to discard tokens that contribute least to the global visual semantics, DivPrune \cite{divprune} and DART \cite{dart} compute pairwise feature similarity to remove the semantically redundant tokens. The second class of methods conducts pruning inside the LLM, utilizing cross-modal attention. For example, FastV \cite{fastv} adopts the attention scores assigned by textual tokens to visual tokens at early LLM layers to remove tokens with low query relevance. SparseVLM \cite{sparsevlm} identifies text tokens that are highly grounded in the image and prunes visual tokens based on the attention they receive from these text tokens. PDrop \cite{pdrop} further introduces a progressive strategy that prunes visual tokens across multiple LLM layers. While pre-LLM methods offer maximal computational savings, their pruning results are text-agnostic. Conversely, intra-LLM methods are text-driven, but pruning at shallow layers suffers from positional bias and attention dispersion, whereas pruning at deeper layers diminishes the efficiency gains by requiring many visual tokens to traverse multiple LLM blocks. These trade-offs motivate the need for an approach that combines the strengths of both categories while mitigating their respective limitations.

\begin{figure}[t]
\centering
\includegraphics[width=1\columnwidth]{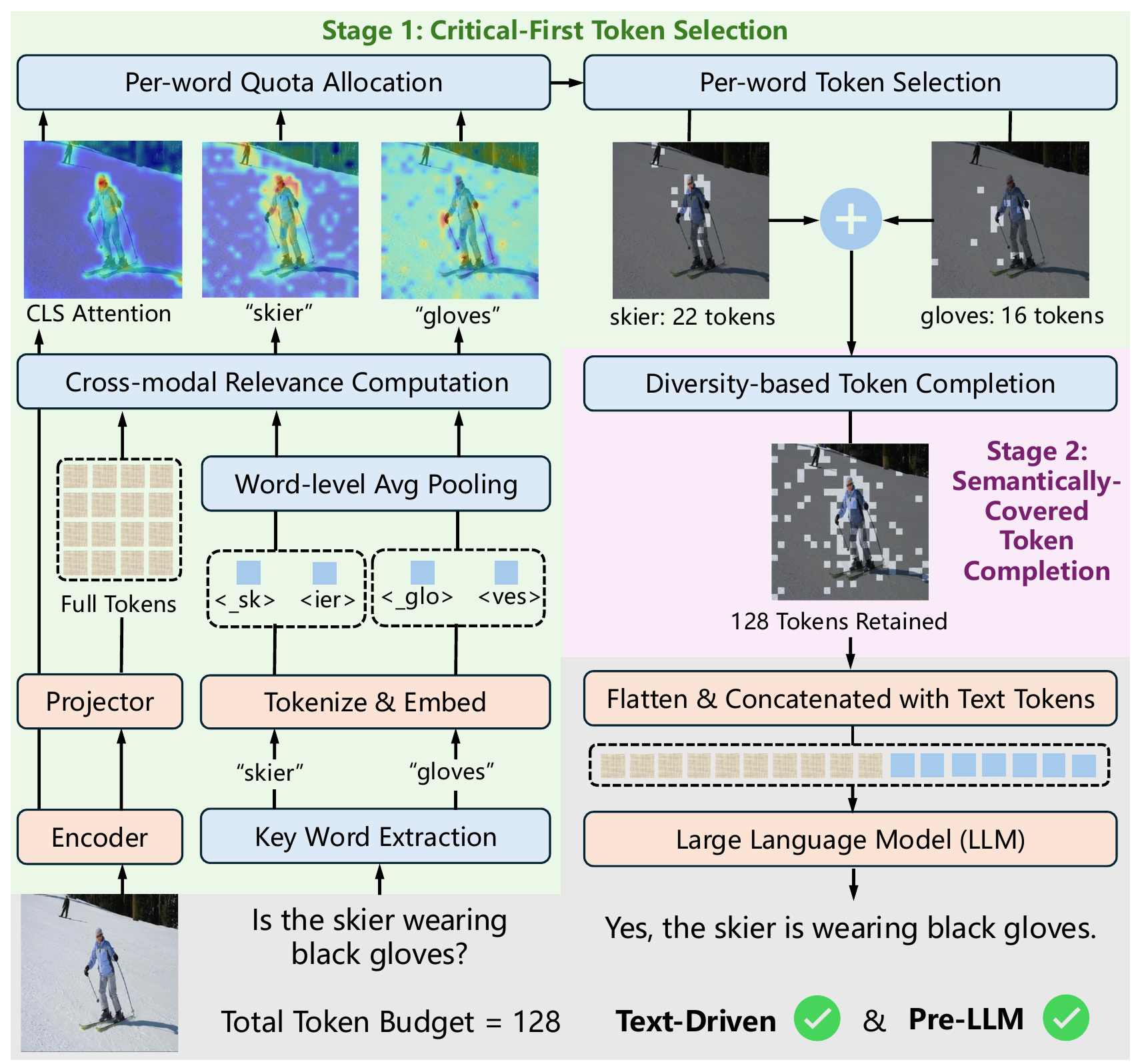}
\caption{Overview of the CRISP workflow. Blue modules represent the components introduced by CRISP, while orange modules denote the original components of the underlying LVLM.}
\label{fig3}
\end{figure}

\section{Method}
\subsection{Overview}
CRISP is a training-free visual token pruning framework that leverages cross-modal signals without entering any Transformer block of the LLM. As illustrated in Fig.~\ref{fig3}, CRISP follows a two-stage pipeline. In the \textit{critical-first} stage, it identifies a compact set of visual tokens that are highly relevant to the key semantic elements of the textual query. In the \textit{semantically-covered} stage, it further augments this subset by selecting diverse and complementary tokens from the remaining pool, thereby preserving essential scene context while maintaining high relevance to the query.

\subsection{Stage 1: Critical-first Token Selection}

Let $\mathbf{V}=\{\mathbf{v}_1, \mathbf{v}_2, \dots, \mathbf{v}_T\}$ denote the projected visual tokens obtained after the projector, where $T$ is the total number of visual tokens. The objective of this stage is to identify a subset $\mathbf{V}_{\mathrm{critical}} \subseteq \mathbf{V}$ that contains the most critical visual cues for answering the current query. To prevent the model from focusing solely on text-aligned evidence while neglecting broader scene context, we impose an upper bound on the number of tokens selected in this stage:
\begin{equation}
\label{eq-alpha}
K_{\mathrm{critical}} = \left\lfloor \alpha \cdot K \right\rfloor,
\end{equation}
where $K$ is the total number of visual tokens to be retained, and $\alpha \in [0,1]$ is a hyperparameter controlling the budget proportion allocated to text-driven selection.

\subsubsection{Key Word Extraction}
As a given query often contains visually irrelevant noise, it is important to extract the content-bearing components of the query. First, we use regular expressions to remove the response instructions (e.g., \textit{``answer the question using a single word or phrase''}). Then, we apply a lightweight algorithm to extract \emph{nouns} from the query, since nouns typically correspond to concrete entities or regions in the image and thus provide the strongest grounding cues. Let the extracted words be $\{w_1, \dots, w_N\}$. As each word may be split into multiple subword tokens, we obtain its word-level embedding by averaging the corresponding token embeddings in the LLM’s input space, resulting in $\{\mathbf{e}_1, \dots, \mathbf{e}_N\}$.

\subsubsection{Cross-modal Relevance Computation}
Since both the projected visual tokens and the word embeddings lie in the same semantic space, we compute the cosine similarity between them to obtain a relevance matrix $\mathbf{S} \in \mathbb{R}^{T \times N}$, where:
\begin{equation}
\mathbf{S}_{i,j} = \frac{\langle \mathbf{v}_i, \mathbf{e}_j \rangle}{\|\mathbf{v}_i\| \, \|\mathbf{e}_j\|}.
\end{equation}
Using this matrix, we can identify the most relevant visual tokens for each key word.

\subsubsection{Word-level Quota Allocation}
As different words could vary in how strongly they are grounded in the image, we assign different token budgets for each word. Intuitively, if a word corresponds to a region that the vision encoder also considers important, it should receive a larger quota. Thus we rely on the CLS attention from the vision encoder. For the word $w_j$, let $a_j$ denote the sum of CLS attention scores over the top $\lfloor K_{\mathrm{critical}}/N \rfloor$ visual tokens most relevant to it (as indicated by $\mathbf{S}_{:,j}$). We define the quota for word $w_j$ as:
\begin{equation}
K_{\mathrm{critical}}^{(j)}
= 
\left\lfloor 
\frac{a_j}{\sum_{j'=1}^{N} a_{j'}} \cdot K_{\mathrm{critical}}
\right\rfloor.
\end{equation}

\begin{table*}[t]
\centering
\caption{Results on LLaVA-1.5-7B. Bold values represent the best score, while underlined values indicate the second-best score. “AVG.\%’’ denotes the overall performance retention rate compared to the full-token baseline.}
\begin{tabular}{l l cccccccccc}
\rowcolor[HTML]{F2F3F5} 
\textbf{Method} & \textbf{Venue} & \textbf{AVG.\%} & \textbf{MME}  & \textbf{GQA}  & \textbf{SQA} & \textbf{POPE} & \textbf{TextVQA} & \textbf{VizWiz} & \textbf{VQA-v2} & \textbf{MMB} & \textbf{MM-Vet} \\
LLaVA-1.5-7B  &    & 100\%            & 1862          & 61.9          & 69.5           & 85.9          & 58.2             & 50.0            & 78.4            & 64.7            & 31.3            \\
\rowcolor[HTML]{F2F3F5} 
\multicolumn{12}{c}{\cellcolor[HTML]{F2F3F5} Retain 192 visual tokens (pruning ratio \(  \approx 66.7\% \))} \\

ToMe\cite{tome}             & ICLR2023          & 89.9\%  & 1563 & 54.3 & 65.2 & 72.4 & 52.1 & 50.0 & 68.0 & 60.5 & 26.8 \\
FastV\cite{fastv}                     & ECCV2024     & 89.5\%  & 1612 & 52.7 & 67.3 & 64.8 & 52.5 & 50.8 & 67.1 & 61.2 & 27.7 \\
LLaVA-PruMerge\cite{prumerge}           & ICCV2025     & 91.5\%  & 1632 & 54.3 & 67.9 & 71.3 & 54.3 & 50.1 & 70.6 & 59.6 & -    \\
FasterVLM\cite{fastervlm}                & arXiv2024    & 97.4\%  & 1768 & 59.3 & 68.6 & 85.2 & 56.8 & 50.1 & 74.3 & 63.6 & 31.8 \\
SparseVLM\cite{sparsevlm}                & ICML2025     & 96.6\%  & 1721 & 57.6 & 69.1 & 83.6 & 56.1 & 50.5 & 75.6 & 62.5 & 31.5 \\
HiRED \cite{hired}                   & AAAI2025     & 94.6\%  & 1737 & 58.7 & 68.4 & 82.8 & 47.4 & 50.1 & 74.9 & 62.8 & -    \\
DART \cite{dart}                    & EMNLP2025    & \underline{98.6\%} & \textbf{1856} & 58.9 & \textbf{69.8} & 82.8 & \underline{57.4} & 51.1 & 76.7 & \underline{63.6} & 31.5 \\
VisionZIP \cite{visionzip}               & CVPR2025     & 98.2\%  & 1767 & 59.3 & 68.9 & 85.3 & 57.8 & 50.9 & 76.5 & 63.0 & 31.7 \\
DivPrune \cite{divprune}                & CVPR2025     & 97.9\%  & 1751 & \underline{59.8} & 69.2 & \underline{87.1} & 54.8 & \underline{51.2} & \underline{76.8} & 63.0 & \underline{32.0} \\
PDrop \cite{pdrop}                   & CVPR2025     & 96.7\%  & 1766 & 57.1 & 68.8 & 82.3 & 56.1 & 51.1 & 75.1 & 63.2 & 30.5 \\
\textbf{CRISP}    & \textbf{Ours}           & \textbf{99.8\%} & \underline{1797} & \textbf{60.3} & \underline{69.4} & \textbf{87.7} & \textbf{59.7} & \textbf{51.4} & \textbf{77.2} & \textbf{63.8} & \textbf{33.0} \\

\rowcolor[HTML]{F2F3F5} 
\multicolumn{12}{c}{\cellcolor[HTML]{F2F3F5}Retain 128 visual tokens (pruning ratio \( \approx 77.8\% \))} \\

ToMe\cite{tome}       & ICLR2023 & 82.9\% & 1343 & 52.4 & 59.6 & 62.8 & 49.1 & 50.1 & 63.0 & 53.3 & 27.2 \\
FastV\cite{fastv}      & ECCV2024 & 83.9\% & 1490 & 49.6 & 60.2 & 59.6 & 50.6 & 51.3 & 61.8 & 56.1 & 28.1 \\
LLaVA-PruMerge\cite{prumerge} & ICCV2025 & 89.5\% & 1554 & 53.3 & 67.1 & 67.2 & 54.3 & 50.3 & 68.8 & 58.1 & - \\
FasterVLM\cite{fastervlm}  & arXiv2024 & 96.4\% & 1769 & 57.9 & 68.6 & 82.6 & \underline{56.9} & 51.1 & 73.1 & 61.3 & 32.5 \\
TRIM \cite{trim}      & COLING2025 & 96.5\% & 1743 & 58.4 & 68.6 & 85.3 & 52.2 & 51.6 & 75.4 & \textbf{63.0} & 29.9 \\
VisPruner \cite{vispruner}  & ICCV2025 & \underline{97.9\%} & 1762 & 58.2 & 69.1 & 84.6 & 57.0 & 52.7 & 75.8 & 62.7 & \textbf{33.7} \\
SparseVLM \cite{sparsevlm} & ICML2025 & 94.5\% & 1696 & 56.0 & 67.1 & 80.5 & 54.9 & 51.4 & 73.8 & 60.0 & 29.0 \\
HiRED \cite{hired}     & AAAI2025 & 93.2\% & 1710 & 57.2 & 68.1 & 79.8 & 46.1 & 51.3 & 73.4 & 61.5 & - \\
DART \cite{dart}      & EMNLP2025 & 97.0\% & \textbf{1845} & 57.9 & \underline{69.1} & 80.1 & 56.4 & 51.7 & 75.9 & 60.7 & 30.9 \\
VisionZIP \cite{visionzip} & CVPR2025 & 97.1\% & 1762 & 57.6 & 68.7 & 83.3 & 56.9 & 51.6 & 75.6 & 62.1 & 32.6 \\
DivPrune\cite{divprune}   & CVPR2025 & 97.2\% & 1694 & \underline{59.4} & 68.5 & \underline{87.0} & 54.5 & \underline{52.7} & \underline{76.0} & 61.5 & 30.7 \\
PDrop \cite{pdrop}     & CVPR2025 & 94.6\% & 1644 & 56.0 & 68.3 & 82.3 & 55.1 & 51.0 & 72.9 & 61.1 & 30.8 \\
\textbf{CRISP} & \textbf{Ours} & \textbf{99.5\%} & \underline{1794} & \textbf{59.7} & \textbf{69.8} & \textbf{87.8} & \textbf{58.5} & \textbf{52.8} & \textbf{76.4} & \underline{62.8} & \underline{32.9} \\

\rowcolor[HTML]{F2F3F5} 
\multicolumn{12}{c}{\cellcolor[HTML]{F2F3F5}Retain 64 visual tokens (pruning ratio \( \approx 88.9\% \))} \\

ToMe \cite{tome}      & ICLR2023 & 73.8\% & 1138 & 48.6 & 50.0 & 52.5 & 45.3 & 49.8 & 57.1 & 43.7 & 25.8 \\
FastV \cite{fastv}     & ECCV2024 & 74.9\% & 1256 & 46.1 & 51.1 & 48.0 & 47.8 & 50.8 & 55.0 & 48.0 & 26.7 \\
LLaVA-PruMerge \cite{prumerge}& ICCV2025 & 88.2\% & 1549 & 51.9 & 68.1 & 65.3 & 54.0 & 50.1 & 67.4 & 55.3 & - \\
FasterVLM \cite{fastervlm} & arXiv2024 & 94.1\% & 1668 & 55.0 & 69.0 & 76.8 & \underline{58.1} & 51.7 & 69.7 & 60.3 & 31.5 \\
TRIM \cite{trim}      & COLING2025 & 94.4\% & 1680 & 56.6 & 69.0 & \underline{85.9} & 49.7 & 51.1 & 72.4 & \underline{60.9} & 24.8 \\
VisPruner \cite{vispruner} & ICCV2025 & 95.3\% & 1674 & 55.4 & \underline{69.1} & 80.4 & 55.8 & \underline{53.3} & 72.7 & \textbf{61.3} & 31.7 \\
SparseVLM \cite{sparsevlm} & ICML2025 & 88.2\% & 1505 & 52.7 & 62.2 & 75.1 & 51.8 & 50.1 & 68.2 & 56.2 & 24.9 \\
HiRED \cite{hired}     & AAAI2025 & 89.5\% & 1599 & 54.6 & 68.2 & 73.6 & 44.2 & 50.2 & 69.7 & 60.2 & - \\
DART \cite{dart}     & EMNLP2025 & 94.3\% & \textbf{1765} & 55.9 & \textbf{69.8} & 73.9 & 54.4 & 51.6 & 72.4 & 60.6 & 26.5 \\
VisionZIP \cite{visionzip}  & CVPR2025 & 94.4\% & 1690 & 55.1 & 69.0 & 77.0 & 55.5 & 52.9 & 72.4 & 60.1 & \underline{31.7} \\
DivPrune \cite{divprune}  & CVPR2025 & \underline{95.7\%} & 1657 & \underline{57.9} & 67.9 & 85.6 & 52.9 & \textbf{53.6} & \underline{74.1} & 60.2 & 29.4 \\
PDrop \cite{pdrop}     & CVPR2025 & 76.3\% & 1092 & 41.9 & 68.6 & 55.9 & 45.9 & 50.7 & 69.2 & 33.3 & 30.7 \\
\textbf{CRISP} & \textbf{Ours} & \textbf{97.7\%} & \underline{1721} & \textbf{58.3} & 68.7 & \textbf{87.2} & \textbf{58.3} & 52.3 & \textbf{75.0} & 60.7 & \textbf{31.8} \\
\end{tabular}
\label{table-1.5}
\end{table*}

\begin{table*}[ht]
\centering
\caption{Results on LLaVA-NeXT-7B. Bold values represent the best score, while underlined values indicate the second-best score. “AVG.\%’’ denotes the overall performance retention rate compared to the full-token baseline.}
\begin{tabular}{l l ccccccccc}
\rowcolor[HTML]{F2F3F5} 
\textbf{Method}  & \textbf{Venue} & \textbf{AVG.\%} & \textbf{MME}  & \textbf{GQA}  & \textbf{SQA} & \textbf{POPE} & \textbf{TextVQA} & \textbf{VizWiz} & \textbf{VQA-v2} & \textbf{MMB} \\

LLaVA-NeXT-7B &  & 100\% & 1851 & 64.2 & 70.1 & 86.5 & 64.9 & 57.6 & 81.8 & 67.4 \\

\rowcolor[HTML]{F2F3F5} 
\multicolumn{11}{c}{\cellcolor[HTML]{F2F3F5}Retain 320 visual tokens (pruning ratio \( \approx 88.9\% \))} \\

FastV\cite{fastv} & ECCV2024 & 88.3\% & 1661 & 55.9 & 62.8 & 71.7 & 55.7 & 53.1 & 71.9 & 61.6 \\
MustDrop\cite{mustdrop} & arXiv2024 & 92.4\% & 1641 & 57.3 & 68.0 & 82.1 & \textbf{59.9} & 54.0 & 73.7 & 62.8 \\
LLaVA-PruMerge\cite{prumerge} & ICCV2025 & 84.9\% & 1534 & 53.6 & 66.4 & 60.8 & 50.6 & 54.0 & 69.7 & 61.3 \\
VisPruner\cite{vispruner} & ICCV2025 & 91.5\% & 1701 & 56.9 & 66.5 & 83.6 & 56.5 & 52.6 & 74.0 & 61.6 \\
SparseVLM\cite{sparsevlm} & ICML2025 & 89.7\% & 1533 & 56.1 & 66.1 & 82.4 & 58.4 & 52.0 & 71.5 & 60.6 \\
HiRED\cite{hired} & AAAI2025 & 93.5\% & 1690 & 59.3 & 66.7 & 83.3 & 58.8 & 54.2 & 75.7 & 64.2 \\
DART\cite{dart} & EMNLP2025 & \underline{95.7\%} & 1726 & \underline{61.7} & \underline{68.4} & 84.1 & 58.7 & 56.1 & \underline{79.1} & \textbf{65.3} \\
VisionZIP\cite{visionzip} & CVPR2025 & 93.9\% & 1720 & 59.3 & 67.3 & 82.3 & 58.8 & \underline{56.2} & 76.2 & 63.1 \\
PDrop\cite{pdrop} & CVPR2025 & 90.7\% & 1663 & 56.4 & 67.5 & 77.6 & 54.4 & 54.1 & 73.5 & 63.4 \\
DivPrune\cite{divprune} & CVPR2025 & 94.4\% & \underline{1731} & 61.1 & 67.7 & \underline{84.7} & 56.2 & 55.6 & 77.2 & 63.9 \\
\textbf{CRISP} & \textbf{Ours} & \textbf{96.9\%} & \textbf{1785} & \textbf{62.2} & \textbf{69.3} & \textbf{88.2} & \underline{58.9} & \textbf{56.3} & \textbf{79.3} & \underline{64.5} \\
\end{tabular}
\label{table-next}
\end{table*}

\subsubsection{Critical Token Selection}
With the per-word quotas determined, we select for each word $w_j$ the top-$K_{\mathrm{critical}}^{(j)}$ visual tokens under the relevance map $\mathbf{S}_{:,j}$. Let the selected set for word $w_j$ be $\mathbf{V}_{\mathrm{critical}}^{(j)}$. The union across all words yields the final critical token set:
\begin{equation}
\mathbf{V}_{\mathrm{critical}} = \mathbf{V}_{\mathrm{critical}}^{(1)}\, \cup \,\mathbf{V}_{\mathrm{critical}}^{(2)}\, \cup \, \dots \, \cup \, \mathbf{V}_{\mathrm{critical}}^{(N)}.
\end{equation}
In rare cases where the noun extraction returns empty (e.g., for queries like \textit{``What is it?''}), CRISP falls back to a purely visual criterion and selects the top-$K_{\mathrm{critical}}$ tokens ranked by CLS attention. Since CLS attention reflects global semantic importance from the vision encoder, this strategy provides a reliable safety mechanism when text guidance is unavailable.

\subsection{Stage 2: Semantically-covered Token Completion}
While the first stage identifies tokens that are strongly aligned with the textual query, relying solely on text-driven evidence risks overlooking important contextual cues (e.g., spatial layout, object relationships, or background semantics). To address this limitation, the second stage of CRISP enriches $\mathbf{V}_{\mathrm{critical}}$ by iteratively selecting the most semantically complementary visual tokens from the remaining pool. 

Let $\mathbf{C}$ denote the current set of selected tokens, which is initialized as $\mathbf{C} = \mathbf{V}_{\mathrm{critical}}$. In each iteration, we compute the maximum similarity to the existing selected tokens for each remaining token $\mathbf{v}_i \in \mathbf{V} \setminus \mathbf{C}$:
\begin{equation}
d_i = \max_{\mathbf{v}_j \in \mathbf{C}} 
\frac{\langle \mathbf{v}_i, \mathbf{v}_j \rangle}{\|\mathbf{v}_i\|\,\|\mathbf{v}_j\|}.
\end{equation}
A small value of $d_i$ indicates that $\mathbf{v}_i$ is semantically dissimilar from the tokens already in $\mathbf{C}$. Therefore, we add the token achieving the lowest maximum similarity to $\mathbf{C}$ and update the similarity scores accordingly. This process continues until the total number of preserved tokens reaches $K$.

This two-stage pipeline operates entirely between the projector and the LLM, requires no additional training or access to the LLM's Transformer block, and can be seamlessly plugged into existing LVLMs as a pre-LLM token pruning module.

\section{Experiments}

\subsection{Experimental Settings}

\subsubsection{Implementation Details}
We validate the effectiveness of CRISP on two representative LVLMs: LLaVA-1.5-7B \cite{llava1.5} and LLaVA-NeXT-7B \cite{llavanext}. Their main difference is that LLaVA-NeXT adopts a multi-crop strategy for high-resolution inputs. For noun extraction within CRISP, we use the 12MB lightweight NLP model \texttt{en\_core\_web\_sm} \cite{en_core_web_sm}. For the hyperparameter $\alpha$ in Eq.~\ref{eq-alpha}, we set it to 0.3 by default. However, when the total token budget is below 100, we increase it to 0.5 to ensure that a sufficient number of text-relevant tokens are retained.

\subsubsection{Benchmarks}
We adopt nine representative benchmarks: VQA-v2 \cite{vqav2}, VizWiz\cite{vizwiz}, and GQA\cite{gqa} for general visual understanding; SQA\cite{sqa} for multimodal knowledge reasoning; TextVQA\cite{textvqa} for text recognition; POPE\cite{pope} for object hallucination assessment; MME\cite{mme}, MMB\cite{mmb}, and MM-Vet\cite{mmvet} for LVLM-specific comprehensive ability evaluation.

\subsubsection{Comparison Methods}
We compare CRISP with representative pruning baselines from both conference and arXiv papers. All baselines are evaluated under matched token budgets. For LLaVA-1.5-7B, we report results with token budgets of 192, 128, and 64 respectively. For LLaVA-NeXT-7B, we evaluate all methods with a token budget of 320.

\subsection{Main Results}
As shown in Table~\ref{table-1.5}, CRISP consistently achieves the strongest overall performance retention on LLaVA-1.5-7B across all pruning budgets. Under a moderate pruning ratio of 66.7\%, CRISP attains an averaged performance of 99.8\%, essentially matching the full-token model and outperforming all the other pruning methods by a clear margin. When the pruning ratio increases to 77.8\%, CRISP still preserves 99.5\% of the full-token performance, whereas the next-best method drops to 97.9\%. Even under the extremely aggressive 88.9\% pruning ratio, CRISP maintains a high performance retention rate of 97.7\%, exceeding the strongest baseline by 2 percentage points.
Beyond averaged results, CRISP demonstrates surprisingly strong performance on POPE, where it even surpasses the full-token baseline with only 64 visual tokens. POPE focuses on object-existence hallucination detection, in which each query explicitly names an object and requires the model to verify whether it appears in the image. This task heavily relies on preserving the few visual regions directly tied to the queried noun. CRISP’s critical-first stage excels in such scenarios by explicitly identifying and retaining the tokens most relevant to the object mentioned in the question, while its semantically-covered stage helps suppress spurious correlations by maintaining contextual diversity. 

The comparison results on LLaVA-NeXT-7B are presented in Table~\ref{table-next}. Under the aggressive 88.9\% pruning ratio, CRISP again delivers the strongest overall performance retention. With only 320 visual tokens retained, it achieves a 96.9\% performance retention rate, surpassing all competing approaches. Also, CRISP obtains the highest accuracy on MME, GQA, SQA, POPE, VizWiz, and VQA-v2, while also remaining highly competitive on TextVQA and MMB. Notably, its POPE score of 88.2 even exceeds that of the full-token baseline (86.5), underscoring CRISP’s ability to preserve object-relevant evidence and effectively suppress hallucinations. These results confirm that CRISP transfers well to more advanced LVLM architectures and maintains strong robustness under extreme token compression.

\subsection{Ablation Study}
To assess the contribution of each main component in CRISP, we conduct ablation experiments under a 128-token budget on LLaVA-1.5-7B. As shown in the upper section of Table~\ref{table-ablation}, removing any component consistently leads to performance drops across the benchmarks. Eliminating the quota allocation mechanism (i.e., assigning uniform token budgets across words) results in noticeable decreases in SQA and MMB scores, indicating that dynamically allocating budgets according to cross-modal grounding strength is crucial. Removing the entire critical-first stage (i.e., relying solely on the semantically-covered mechanism) causes a more substantial degradation, particularly in MME, confirming that text-driven selection is essential for preserving instruction-relevant visual information. Similarly, removing the semantically-covered stage and relying solely on text relevance also leads to clear declines across all benchmarks, especially on MMB, demonstrating that diversity-based coverage is necessary for maintaining sufficient contextual grounding. Together, these results validate both the effectiveness and the complementary roles of the two stages in CRISP’s design.

The lower section of Table~\ref{table-ablation} analyzes the impact of varying $\alpha$, which controls the proportion of the token budget allocated to the critical-first stage. Small values of $\alpha$ underweight textual relevance, causing the model to depend too heavily on feature diversity, whereas large values strengthen text-driven selection at the cost of contextual completeness. The results show that our default setting of $\alpha = 0.3$ provides the best overall performance across benchmarks. This highlights a key insight: in most cases, only a modest number of text-relevant tokens is needed to preserve critical visual evidence, while maintaining adequate contextual cues is essential for robust and comprehensive visual understanding.

%%=========================================
\begin{table}[t]
\centering
\caption{Ablation study of CRISP on LLaVA-1.5-7B under a 128-token budget.}
\begin{tabular}{lccccc}
\rowcolor[HTML]{F2F3F5} 
                       & MME           & GQA           & SQA           & POPE          & MMB        \\
\rowcolor[HTML]{FFF5EB} 
Full CRISP                   & \textbf{1794} & \textbf{59.7} & \textbf{69.8} & \textbf{87.8} & \textbf{62.8} \\
\rowcolor[HTML]{FFF5EB} 
w/o quota allocation   & 1759          & 59.5          & 68.7          & 87.8          & 61.9          \\
\rowcolor[HTML]{FFF5EB} 
w/o stage-1 selection  & 1694          & 59.3          & 68.5          & 87.0          & 61.4          \\
\rowcolor[HTML]{FFF5EB} 
w/o stage-2 completion & 1738          & 58.9          & 68.3          & 86.8          & 59.8          \\
\rowcolor[HTML]{F0FBEF} 
$\alpha$ = 0.1            & 1709          & 59.4          & 69.0          & 87.6          & 61.8          \\
\rowcolor[HTML]{F0FBEF} 
$\alpha$ = 0.3            & \textbf{1794} & \textbf{59.7} & \textbf{69.8} & \textbf{87.8} & \textbf{62.8} \\
\rowcolor[HTML]{F0FBEF} 
$\alpha$ = 0.5            & 1779          & 59.5          & 69.2          & 87.7          & 62.0          \\
\rowcolor[HTML]{F0FBEF} 
$\alpha$ = 0.7            & 1777          & 59.6          & 68.8          & 87.2          & 61.3          \\
\rowcolor[HTML]{F0FBEF} 
$\alpha$ = 0.9            & 1775          & 59.3          & 68.7          & 87.1          & 61.4         
\end{tabular}
\label{table-ablation}
\end{table}

%%==========================================

%%=========================================
\begin{table}[t]
\centering
\caption{Efficiency analysis of CRISP with LLaVA-1.5-7B and LLaVA-NeXT-7B on the MME benchmark.}
\setlength{\tabcolsep}{2pt}
\begin{tabular}{lccccc}
\rowcolor[HTML]{F2F3F5} 
Method    & \# Tokens & \begin{tabular}[c]{@{}c@{}}FLOPs\\ (T)\end{tabular} 
          & \begin{tabular}[c]{@{}c@{}}KV-cache\\ (MB)\end{tabular} 
          & \begin{tabular}[c]{@{}c@{}}Decoding Speed\\ (tokens/sec)\end{tabular} 
          & \begin{tabular}[c]{@{}c@{}}Total Time\\ (min:sec)\end{tabular} \\

\rowcolor[HTML]{FFF5EB}
LLaVA-1.5-7B   & 576  & 8.47  & 318.6 & 22.0 & 2:21 \\
\rowcolor[HTML]{FFF5EB}
+ CRISP        & 128  & 2.46  & 94.6  & 45.0 & 1:35 \\
\rowcolor[HTML]{FFF5EB}
+ CRISP        & 64   & 1.63  & 62.6  & 48.1 & 1:23 \\

\rowcolor[HTML]{F0FBEF}
LLaVA-NeXT-7B  & 2880 & 30.57 & 1084.7 & 18.4 & 4:29 \\
\rowcolor[HTML]{F0FBEF}
+ CRISP        & 640  & 9.34  & 350.6  & 37.8 & 2:24 \\
\rowcolor[HTML]{F0FBEF}
+ CRISP        & 320  & 5.01  & 190.6  & 40.4 & 1:56 \\

\end{tabular}
\label{table-efficiency}
\end{table}

%%============================

\subsection{Practical Efficiency Analysis}
We further evaluate the practical efficiency benefits brought by CRISP. We report four metrics: (1) FLOPs, measuring the average computational cost per sample; (2) KV-cache memory, representing the average key–value cache size per sample; (3) decoding speed, measuring how many tokens are generated per second; and (4) total time, denoting the total time used to complete the full benchmark evaluation.

As shown in Table~\ref{table-efficiency}, CRISP consistently reduces computational overhead and substantially accelerates inference across both models. Its pre-LLM pruning nature yields near-linear reductions in FLOPs and KV-cache usage as the number of retained visual tokens decreases. These improvements directly translate into faster decoding: CRISP achieves more than a 2$\times$
speedup on both models. The end-to-end evaluation time is also significantly shortened. Together with the strong performance retention rates reported in Tables~\ref{table-1.5} and~\ref{table-next}, it can be confirmed that CRISP provides an excellent trade-off between accuracy and efficiency.

\section{Conclusion}
We introduced CRISP, a training-free visual token pruning framework that performs text-driven token selection entirely before the LLM. Extensive experiments across multiple LVLMs and pruning settings show that CRISP consistently achieves strong performance retention while significantly reducing inference cost. These results highlight CRISP as a practical and generalizable solution for accelerating LVLMs, particularly in latency-sensitive deployment scenarios.

\section*{Acknowledgment}

This work was supported by the Lingang Laboratory (Grant No.LGL-1987-10).

\bibliographystyle{IEEEtran}
\bibliography{refs}

\end{document}